\documentclass{article}

\usepackage[nonatbib, final]{neurips_2024}

\usepackage[utf8]{inputenc} 
\usepackage[T1]{fontenc}    
\usepackage{hyperref}       
\usepackage{url}            
\usepackage{booktabs}       
\usepackage{amsfonts}       
\usepackage{nicefrac}       
\usepackage{microtype}      
\usepackage{xcolor}         
\usepackage{graphicx}
\usepackage{amsmath}
\usepackage{animate}
\usepackage{media9}

\title{Towards Kinetic Manipulation of the Latent Space}

\author{%
  Diego Porres\\
Computer Vision Center (CVC) \& Universitat Aut\`onoma de Barcelona\\
  Barcelona, Spain, 08193 \\
  \texttt{dporres@cvc.uab.es} \\
}

\begin{document}

\maketitle

\begin{abstract}

The latent space of many generative models are rich in unexplored valleys and mountains. The majority of tools used for exploring them are so far limited to Graphical User Interfaces (GUIs). While specialized hardware can be used for this task, we show that a simple feature extraction of pre-trained Convolutional Neural Networks (CNNs) from a live RGB camera feed does a very good job at manipulating the latent space with simple changes in the scene, with vast room for improvement. We name this new paradigm \textit{Visual-reactive Interpolation}, and the full code can be found at \href{https://github.com/PDillis/stylegan3-fun}{https://github.com/PDillis/stylegan3-fun}.
\end{abstract}

\section{Motivation and Background}

Interaction with generative models such as Generative Adversarial Networks (GANs) \cite{GAN} and diffusion-based text-to-image models (T2I) \cite{diffusion} has been relinquished to the land of software, specifically Graphical User Interfaces (GUIs). The main advantage of using these is the speed at which they can be modified and shared between users, in addition to simplifying interaction with the model, including it's high-dimensional latent space. However, we believe this also introduces a rift between the generative models and their users, if not also with their creators, as we forego the most basic interface: the human body.
	
Indeed, \cite{expressive_body} notes, among other things: firstly, human beings possess a 'centric' quality (they interact with their environment from the vantage point of a \textit{centre}); secondly, we shouldn't separate human bodies from the minds, but as acting mind-body unities; and thirdly, the body is instrumental in communication both within the self as with the environment (be it motor or hormonal activity). Our goal is to create a system by which we can have a live performer be the one controlling the latent space of a generative model with their body or facial movements and scenery changes via moving objects in the scene.

We argue then, that there are two \textit{centers} on an image, whether real or synthesized: the subject or focus of the image, and the camera (real or virtual). As such, we should then aim at letting the camera controller to be another actor in the image synthesizing pipeline by changing the focused area of the performer, lightning, or even camera lenses. For our purposes, we wish the two actors to interact through a manipulation of the latent space, made visible via the synthesized images that a pre-trained Generator of a GAN will provide. This will allow the two to forge a common story. This work aims to document the creation of this tool.

\begin{figure}[ht]
    \centering
    \animategraphics[label=demo,loop,width=0.8\linewidth]{20}{images/vr_sg2_frames/out}{001}{141}
    \newline
    \mediabutton[jsaction={
    	if(anim['demo'].isPlaying)
    	anim['demo'].pause();
    	else
    	anim['demo'].playFwd();
    }]{\fbox{Play/Pause}}
    \caption{\textbf{Test 1:} Visual-reactive live demo using a StyleGAN2 model ($512\times512$ resolution) trained on urban scenes from the A2D2 dataset \cite{geyer2020a2d2}. We use style mixing with a static latent, and the encoded camera image (resolution $426\times320$) will procure the \texttt{coarse} and \texttt{middle} noise scales. Note the video control above does not work in browsers, but works fine with Adobe Acrobat. 
    Click \href{https://drive.google.com/file/d/1BrC3PulFpdtBdM6p97MaGN43jU4EOBoo/view?usp=sharing}{here} for an online version of the video.}
    \label{fig:demo_animation}
\end{figure}

\section{Latent Space Interaction}\label{latent_interaction}
	
Since our aim is to control the image synthesis generation during a live performance show, we will quickly explore other already existing alternatives. Concretely, we will focus our attention to manipulate the latent space of StyleGAN1/2/2-ADA \cite{SGAN, SGAN2, SGAN2-ADA}, some of the most popular and widely trained networks, but we will look at inspiration for what has been done in others as well. Our focus is on GANs as as their sampling cost is low compared to other models, which translates to a smoother live performance, but hopefully this work (or an iteration of it) can be also applied to other generative model such as T2I latent diffusion models \cite{latentdiffusion}.

There have been many recent efforts towards enhancing the interaction with generative models, both offline as well as during live shows. Most notably, \cite{maua_2020} uses and exploits the architecture of StyleGAN1/2 (noise injection, network bending \cite{network_bending}, style mixing, among others), as well as the features of music itself for creating enthralling audio-reactive latent interpolations. This allows for expressive coverage of the latent space and sets the stage for further expanding the our pipeline's capacities.

\cite{livegan} questions the interactive capabilities of a mouse and keyboard when trying to explore interactions with trained GANs. A such, they program a MIDI controller to guide StyleGAN1/2/2-ADA, selecting different controllers to change local or global variables, such as the truncation factor $\psi$ or for quick sampling of the latent space. While this requires adding hardware and its respective programming, it highlights the general need for better tools and methods of the to explore the latent space of generative models. Likewise, text (or speech) can be used to interact with generative models on how to edit the synthesized 2D image \cite{interactivemlgen2023} or 3D animation \cite{huang2024realtimeanimationgenerationcontrol}.

On the other hand, \cite{gans_interactive} seeks to project a live video feed into the disentangled latent space $\mathcal{W}$ of StyleGAN. A camera feed is iteratively projected to $\mathcal{W}$ and then used to synthesize an image, showcasing the expressivity of the network. However, we must caution that this process has a model selection bias, as FFHQ's latent space is perhaps the only model where we can effectively project any image \cite{img2sgan}.
    
Independently of live shows, tools such as GANSpace \cite{GANspace} allow for automatically creating interpretable controls in the latent space of StyleGAN or BigGAN \cite{bigGAN} by identifying important latent directions using PCA. This method is limited to static sliders that are not always the directions the user wishes to move towards, but we note that these could be mapped to different actions in the scene. DragGAN \cite{draggan} and DragDiffusion \cite{shi2023dragdiffusion} seek to give more precise editing control to the user by adding points to "drag". However, the movement of these points is an optimization, increasing the compute cost of this approach.



For real-time generation, Xoromancy \cite{Crawford2019} has explored hand-based latent space manipulation. However, it relies on specialized hardware (\href{https://www.ultraleap.com/}{Leap Motion}) and fixed camera placement, limiting its accessibility and flexibility. In the same vein, the art installation \textit{Fencing Hallucination} \cite{Qiu2023CombatingT} uses a \href{https://en.wikipedia.org/wiki/Kinect}{Kinect} to accurately extract the pose of the user as part of its pipeline. We argue that it is possible to eliminate the need for specialized equipment by using readily available pre-trained models and affordable cameras, in turn broadening the scope of interaction with generative models and democratizing the access to latent space manipulation.

Furthermore, by removing the constraint of fixed camera positioning, we introduce the camera itself as a "second actor" in the image generation pipeline, opening up new possibilities for performative art where both the subject and the camera operator can collaboratively influence the generated output, effectively turning the computer as a \emph{peer} and not just as a \emph{sub-contractor} \cite{lin2023ontologycocreativeaisystems}. The flexibility of our approach extends to its compatibility with a wide range of pre-trained GANs and in the future to Text-to-Image (T2I) Diffusion Models \cite{diffusion, latentdiffusion} or Consistency Models \cite{Song2023ConsistencyM, Luo2023LatentCM}, enhancing its applicability and potential for creative expression. 

\section{Visual-reactive Interpolation}\label{visual_interp}
	
Given that StyleGAN lacks an encoder back into its latent space $\mathcal{Z}$, unlike other architectures such as BiGAN \cite{BiGAN}, we must look at alternatives. Although using the last Fully Connected layer of the GAN's Discriminator has been proposed for this purpose as it has the desired dimensionality and is essentially the same network \cite{disc_synth}, our tests have shown that the resulting encoding fails to cover the majority of the latent space, resulting in nigh-static images.

Training an encoder for each available pre-trained GAN is out of the question, so we opt to use pre-trained feature extractors $F$. As such, we will use the intermediate representations of these, effectively acting as encoders into the latent space $\mathcal{Z}$. We discard using SOTA models such as CLIP \cite{CLIP}, as we wish to minimize the footprint of this part of the model as much as possible. Likewise, Vision Transformers (ViT) \cite{ViT} have shown great promise in image classification tasks, but our experiments show that almost no benefit is seen when used in our pipeline. For now, we gear towards VGG16 \cite{vgg16}, as its intermediate representations have some desired characteristics, but note that future work can make use of more recent architectures that are lighter to run on the edge.

We start our testing by exploiting these feature extractors as well as the style-mixing characteristic of StyleGAN. Based on user feedback, we then move to more fine-grained control of specific parts of the Generator. Lastly, we list some possible venues to take, but note that the possibilities are endless, especially if we include multimodal inputs to our setting.

\subsection{Test 1: Visual Encoding and Style Mixing}\label{subsec:test1}

In the following, we will use the notation of \cite{styletransfer_gatys}. We denote by $x$ the frame captured by a camera $C$ that we will feed to the selected feature extractor $F$. When we pass each frame through this network, it will be encoded at layer $l$ as $F^{l}\in\mathbb{R}^{N^l \times M^l}$,where $N^{l}$ denotes the number of feature maps of size $M^{l}$. The size refers to the product of the height $h^{l}$ and width $w^{l}$ of the feature maps, so $M^{l}=h^{l}\cdot w^{l}$, with $h^{l},w^{l}\in\mathbb{Z}^{+}$. We must convert these feature maps into a latent, for which we make use of a function $g$. For more details, see Appendix \ref{ap:feature}.

\begin{figure}[ht]
    \centering
    \includegraphics[width=\linewidth]{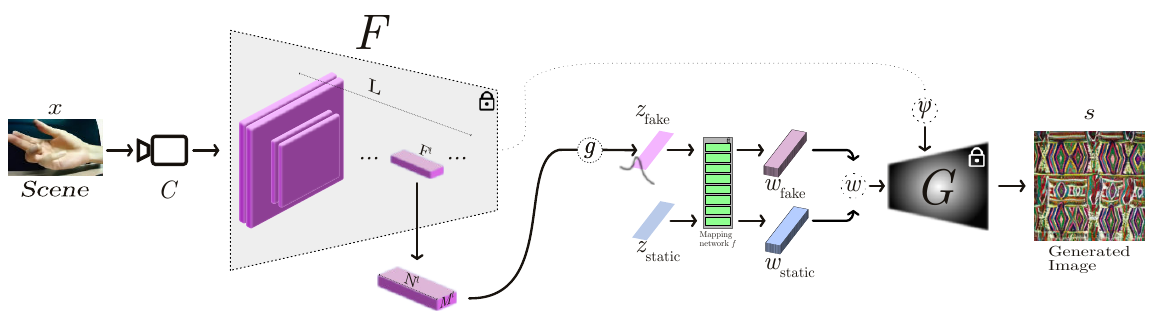}
    \caption{\textbf{Test 1 pipeline.}
    A frame $x$ of a scene is captured with a camera $C$, which is then fed to the feature extractor $F$. We select $F$ to contain $L$ convolutional layers. At layer $l$, the representation of $x$ will be $F^{l}$. We pass this representation through our selected function $g$, turning it into a latent vector $z_{\text{fake}}$, which is then passed to the (frozen) mapping network $f$ of the Generator. To perform style-mixing, we can use a static latent $z_{\text{static}}$, to produce a final disentangled latent vector $w$. It, along the truncation parameter $\psi$, will be used by $G$ to generate the final synthesized image $s$. Note that $\psi$ could also be influenced or controlled via the encoded scene.}
    \label{fig:pipeline}
\end{figure}

The dimensions of $x$ must also be carefully selected, as this will also affect the real-time inference that can be achieved by the available compute. Note that the camera $C$ is not limited to being an RGB camera. Indeed, this technique works with any other type of sensor, so long as a pre-trained network $F$ is available (or equivalent data pipeline). The full pipeline for this process is shown in Figure \ref{fig:pipeline} and a demo is shown in Figure \ref{fig:demo_animation}. The input image has been encoded to follow a strict $4:3$ aspect ratio, but this can be modified. 

\paragraph{User Feedback} Initial user testing at the \href{https://www.cvc.uab.es/experimentai/sessio-3-art-i-ia/}{\emph{ExperimentAI 2023}} and the \href{https://img.beteve.cat/wp-content/uploads/2023/06/Programa_Festa_de_la_ciencia.pdf}{\emph{Festa de la Ci\`encia 2023}} revealed that users were intrigued by this new interaction method, but they quickly became disengaged due to the lack of fine-grained control. Users found that changing objects in the scene sometimes resulted in camera autofocusing, reverting the generated image to its original state. This feedback prompted us to explore more precise manipulation techniques in our subsequent tests.

\subsection{Test 2: Manipulation of Learned Constants}\label{subsec:test2}

Based on user feedback from Test 1 (Section \ref{subsec:test1})  indicating a need for more fine-grained manipulation, we turned our attention to manipulating specific parts of the Generator, particularly some of the learned constants. During training, StyleGAN2 learns a constant parameter (\texttt{G.synthesis.b4.const}) that will aid it in positioning certain aspects of the generated image (eyes, nose, ears, hairs when tasked with mimicking faces, for example). If we corrupt this, then the network will lose track of where to place the eyes, for example, in turn generating multiple pairs of them. This is essentially what \href{https://github.com/aydao/stylegan2-surgery}{Flesh Digressions} does, which we adapt to our setting.

A simple way to corrupt this constant is by extracting keypoints from one or more human body parts using \href{https://github.com/google-ai-edge/mediapipe}{MediaPipe} or \href{https://github.com/open-mmlab/mmpose}{MMpose}. We can define the center of the image as the learned constant and the average distnace of the selected body part will define the amount of corruption we introduce to this constant. We show a demo in Figure \ref{fig:demo_sg2_flesh}.

On the other hand, StyleGAN3 \cite{SGAN3} learns a fixed affine transform to apply to the latent space in order to correctly translate (and optionally rotate) the generated objects. This affine transform can also be corrupted in a more intuitive way: for example, by calculating the angle between the vertical axis and the middle finger, we can set how much to rotate the image. The distance form the hand to the center of the image can set how much to scale the image. We showcase a demo in Figure \ref{fig:demo_sg3_affine}.

\begin{figure}[ht]
    \centering
    \animategraphics[label=demo2,loop,width=0.8\linewidth]{30}{images/vr_sg2_fleshdig_frames/out}{271}{450}
    \newline
    \mediabutton[jsaction={
    	if(anim['demo2'].isPlaying)
    	anim['demo2'].pause();
    	else
    	anim['demo2'].playFwd();
    }]{\fbox{Play/Pause}}
    \caption{\textbf{Test 2:} Live manipulation of the learned constant in StyleGAN2. Note the video control above does not work in browsers, but works fine with Adobe Acrobat.
    Click \href{https://drive.google.com/file/d/1mKrq7Q0CSoQRqBcONl1DttJwgyKIs-Ls/view?usp=sharing}{here} for an online version of the video.}
    \label{fig:demo_sg2_flesh}
\end{figure}

\begin{figure}[ht]
    \centering
    \animategraphics[label=demo3,loop,width=\linewidth]{30}{images/vr_sg3_frames/out}{001}{180}
    \newline
    \mediabutton[jsaction={
    	if(anim['demo3'].isPlaying)
    	anim['demo3'].pause();
    	else
    	anim['demo3'].playFwd();
    }]{\fbox{Play/Pause}}
    \caption{\textbf{Test 2:} Manipulating the learned affine transformation matrix in StyleGAN3. Note the video control above does not work in browsers, but works fine with Adobe Acrobat. Click \href{https://drive.google.com/file/d/1msCAXIIYHU4u_uAjpsJCTzYdcEygPnQ-/view?usp=sharing}{here} for an online version of the video.}
    \label{fig:demo_sg3_affine}
\end{figure}

\paragraph{User Feedback} Originally, we had intended to use two hands to perform these corruptions. However, user testing at the \href{https://img.beteve.cat/wp-content/uploads/2023/06/Programa_Festa_de_la_ciencia.pdf}{\emph{Festa de la Ci\`encia 2023}} revealed significant challenges with this approach. While the system ran smoothly with one hand, the introduction of a second hand caused confusion. Despite MediaPipe's ability to distinguish between left and right hands, the active hand controlling the movement could unpredictably switch between left and right when both were detected. This led to a counterintuitive experience, disrupting users' spatial understanding and control. To address these issues, we ultimately opted for a single-hand approach. This decision resulted in much cleaner and better results, as well as maintaining users' interest in exploring how to manipulate these systems.

\section{Future Work}\label{ap:future}

While the code is readily available at \url{https://github.com/PDillis/stylegan3-fun}, 
we note that it will be changing constantly as we introduce more options to the user, as well as perform more tests with it. The main bulk of work will be dedicated to port all of the previous tests to the GUI provided in the StyleGAN3 repository. 

Beyond run-time optimization, the following ideas are planned to be explored to some extent in the coming months, or at least variations of them. We note, however, that the proposed parameters to manipulate via visual-reactive interpolation are interchangeable, and even more can be exploited from each model:
	
\begin{itemize}
    \item Replace VGG16 with smaller footprint networks such as MobileNetV3 \cite{mobilenetv3} or EfficientNet-B0 \cite{efficientnet}.
    \item Conversely, use \emph{self-supervised} visual features such as DINOv2 \cite{oquab2023dinov2}.
    \item Use monocular depth estimation or optical flow estimation for manipulating the truncation trick parameter $\psi$ via e.g. the normalized average depth of the scene using pre-trained models such as Depth-Anything-V2-Small \cite{depth_anything_v2}.
    \item $\psi$ can be thought of as controlling the expressiveness of the generator $G$. We can place virtual objects (such as disks) in a scene and allow users to manipulate them with their hands. By calculating the total kinetic energy $K=\sum_{i} m_{i} \|\mathbf{v}_{i}\|^2$, we can dynamically adjust $\psi$ and thus the expressivity of $G$.
    \item Use semantic segmentation or classification models for exploiting class-conditional StyleGAN models, whether for manipulating the latent vectors or the class vectors themselves.
    \item Using PCA to extract notable directions in the latent space, as done in GANSpace \cite{GANspace}, and move towards/away from them using specific facial gestures. 
    \item Add audiorreactive capabilities \cite{maua_2020}, effectively generating \textbf{\textit{audiovisual}}-reactive interpolations, allowing for a live band to join the performance.
    \item Add network bending \cite{network_bending} capabilities to the extracted features, or for specific regions in the image.
    \item Extend this work to T2I Diffusion Models \cite{diffusion, latentdiffusion}, with a focus on Consistency Models \cite{Song2023ConsistencyM, Luo2023LatentCM} for real-time manipulation.
\end{itemize}	



\bibliographystyle{alpha}
\bibliography{references}

\appendix
\section*{Appendix}
\section{Feature extraction}\label{ap:feature}
\paragraph{Transformation into $\mathcal{Z}$} After capturing the internal representations, we wish to convert these into useful objects that the Generator may then use: latent vectors. We are then interested in a family of functions $\mathcal{G}$ that will transform our intermediate representation $F^{l}$ into a vector $z_{\text{fake}}$, i.e., turn the features into \textit{fake} latent vectors to feed our Generator $G$. We carefully select this family of functions such that, 

\begin{equation}\label{eq:functions}
\mathcal{G} \ni g: F^{l} \to \mathcal{Z} \sim \mathcal{N}(0, \mathbb{I})
\end{equation}
	
Afterwards, the mapping function $f$ of StyleGAN (\texttt{G.mapping}) will be in charge of mapping these "fake latents" into the disentangled latent space $\mathcal{W}$. For example, at its most basic, $g$ could simply be the channel-wise average of the feature representation $F^{l}$ of VGG16. More generally, we can also do a weighted average of different layers $l$ (as is done in the content loss in neural style transfer \cite{styletransfer_gatys}), e.g.:

\begin{equation}\label{eq:default_f}
g(F^{l})=\alpha_{l} \sum_{l} \left(\frac{1}{M^{l}} \sum_{i,j}F_{ij}^{l}\right)
\end{equation}

where $F_{ij}^{l}$ denotes the $i$-th filter of layer $l$ at position $j$, and $\alpha_{l}$ is the weight given to the representation at layer $l$. Our experiments so far have shown that using a single layer is sufficient, but leave the possibility to use more to each individual case. 
	
We note this could limit the expressiveness of our generator $G$, so we could ensure the condition in Equation \ref{eq:functions} is met. However, we have found that Equation \ref{eq:default_f} is sufficient, mostly because of the selection of VGG16 and its separation of features in the last layers is channel-wise, which matches the dimensionality of the default latent space $\mathcal{Z}$ of StyleGAN. 

\paragraph{Layer selection}

The default dimensionality of StyleGAN's latent space is $\lvert \mathcal{Z}\rvert=512$. Then, for simplicity, we are interested in the layers that have $N^{l}=\lvert\mathcal{Z}\rvert=512$. Table \ref{table} shows the convolutional layers in VGG16 that share this property, but note that our code is general enough that \textit{any} layer (convolutional or not) in VGG16 can be used to obtain $z_{\text{fake}}$. The user should experiment and select the best option for their particular setup.

\begin{table}[h]
    \caption{Feature representations names and shapes for an $256\times256$ RGB image passed through a VGG16 network. We showcase here only those for which $N^{l}=|\mathcal{Z}|=512$.}
    \label{table}
    \centering
    \begin{tabular}{lll}
        \toprule
        \textbf{Layer Name}     & $\mathbf{N^{l}}$     & $\mathbf{M^{l}}$ \\
        \midrule
        \texttt{conv4\_1} & 512 & $32\times32$  \\
        \texttt{conv4\_2} & 512 & $32\times32$  \\
        \texttt{conv4\_3} & 512 & $32\times32$  \\
        \midrule
        \texttt{conv5\_1} & 512 & $16\times16$  \\
        \texttt{conv5\_2} & 512 & $16\times16$  \\
        \texttt{conv5\_3} & 512 & $16\times16$  \\
        \midrule
        \texttt{adavgpool} & 512 & $7\times7$\\
        \bottomrule
    \end{tabular}
\end{table}

\paragraph{Image Synthesis} To make better use of the expressivity of the trained StyleGAN, we can partition $F^{l}$ into different sections before passing it into $g$, each producing a different disentangled latent vector $w$. In Figure \ref{fig:demo_animation}, we used $w_{\text{coarse}}$, $w_{\text{middle}}$, and $w_{\text{fine}}$ to refer to the disentangled latent vectors obtained from the top, bottom left, and bottom right parts of the input image $x$. Furthermore, we can set a static latent vector $w_{\text{static}}$ and use the style-mixing properties of StyleGAN to more effectively control different aspects of the synthesized image. For example, one region of the input image will control the fine details (colors) of the synthesized image ($w_{\text{fine}}$), while in others control larger structures ($w_{\text{coarse}}$ and $w_{\text{middle}}$).

\end{document}